# Rise of the Autonomous Machines

Shaoshan Liu[1] *Senior Member IEEE*
Jean-Luc Gaudiot[2] *Life Fellow IEEE*

[1]PerceptIn Inc
[2]University of California, Irvine, U.S.A.

## Abstract:

*After decades of uninterrupted progress and growth, information technology has so evolved that it can be said we are entering the age of autonomous machines, but there exist many roadblocks in the way of making this a reality.  In this article, we make a preliminary attempt at recognizing and categorizing the technical and non-technical challenges of autonomous machines; for each of the ten areas we have identified, we review current status, roadblocks, and potential research directions.  It is hoped that this will help the community define clear, effective, and more formal development goalposts for the future.*

## The Age of Autonomous Machines: The Sixth Layer of Information Technology

As an increasing number and kinds of autonomous machines enter our daily life, the age of autonomous machines is upon us and a whole new era of information technology begins.  How did we get here?  Before delving into autonomous machines, let us first review the evolution of computing machinery.   Information technology took off in the 1960s when Fairchild Semiconductors and Intel established its foundation by producing the first silicon microprocessors, with the attendant explosive growth of the Silicon Valley (See Figure 1).  At first, although microprocessor technologies greatly improved industrial productivity, the general public had limited access to it.  This changed in the 1980s, with the advent of personal computers, and later Apple Macintosh and Microsoft Windows, using a Graphical User Interface (GUI).  The second layer had been laid, and the vision of ubiquitous computers, targeted for use by untrained personnel at home, had become a possibility.

With virtually everyone having access to computing power by the early 2000s, Yahoo and Google laid the third layer, connecting people—indirectly, with information available through search engines. With the third layer, the internet, a core infrastructure to enable later layers, could become ubiquitous.

Beginning with Facebook in 2004, social networking sites created the fourth layer of information technology by allowing people to directly connect with each other, effectively moving the whole of human society to the World Wide Web.

As the population of Internet-savvy people reached critical mass, the emergence of such applications as Airbnb (2008), or Uber (2009) was the basis for the fifth layer by providing direct Internet commerce services.

So far, each new layer of information technology, with its added refinements, has incrementally improved popular access and demand. Note that for most Internet commerce sites providing access to service providers through the Internet, it is still ultimately humans who are providing the services.

We have now entered the age of autonomous machines (the sixth layer), autonomous machines, such as service robots, autonomous drones, delivery robots, and autonomous vehicles, rather than humans, will provide services. Obviously, autonomous machines have the potential to completely upend our daily life and our economy in the coming decade.

Yet, autonomous machines are extremely complex systems that integrate many pieces of technologies [1] and for autonomous machines to become an integral part of our daily life, we are still facing many technical and non-technical challenges. In this article, we categorize these challenges in ten areas. For each area, we introduce the current status and roadblocks, as well as potential research directions. It is hoped that this will help the community define clear, effective, and more formal development goalposts for the future.

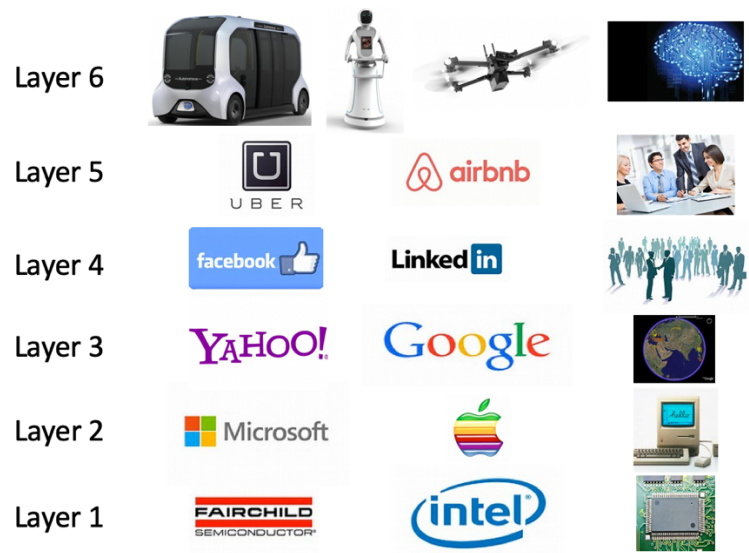

Figure 1: The Layers of Information Technology

## Area 1: The On-Machine Compute System

As opposed to other computing workloads, autonomous machines have a very deep processing pipeline, or computation graph, with strong dependencies between the different stages and strict deadlines associated with each stage [2]. For instance, Figure 2 presents an overview of the processing pipeline of a level 4 autonomous driving system.

Starting from the left side of the figure, the system consumes raw sensing data from mmWave radars, LiDARs, cameras, and Global Navigation Satellite System (GNSS) receivers and Inertial Measurement Units (IMUs), where each sensor produces raw data at its own frequency: the cameras capture images at 30 FPS and feed the raw data to the *2D Perception module*, the LiDARs capture point clouds at 10 FPS and feed the raw data to the *3D Perception module* as well as the *Localization module*, the GNSS/IMUs generate positional updates at 100 Hz and feed the raw data to the *Localization module*, and the mmWave radars detect obstacles at a rate of 10 FPS. All this raw data is then fed to the *Perception Fusion module*.

Next, the results of the *2D* and *3D Perception Modules* are fed into the *Perception Fusion module* at 30 Hz and 10 Hz respectively to create a comprehensive perception list of all detected objects. The perception list is then sent to the *Tracking module* at 10 Hz to create a tracking list of all detected objects. The tracking list then is fed into the *Prediction module* at 10 Hz to create a prediction list of all objects. After that, both the prediction results and the localization results are received by the *Planning module* at 10 Hz to generate a navigation plan which then goes into the *Control module* at 10 Hz to generate control commands. These are in turn sent to the autonomous machine for execution at 100 Hz.

Hence, every 10 ms, the autonomous machine needs to generate a control command to maneuver the autonomous machine. If any upstream module, such as the *Perception module*, misses the deadline to generate an output, the *Control module* must still generate a command before the deadline. This could lead to disastrous results as the autonomous machine would then be essentially driving blindly without timely participation from the perception unit.

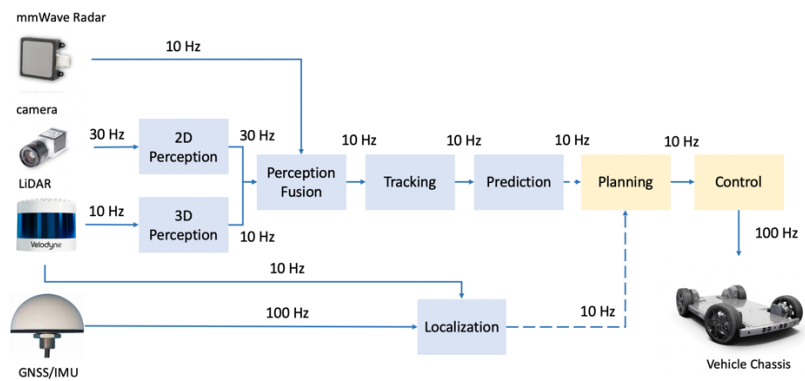

Figure 2: The processing pipeline of an autonomous vehicle

To minimize the end-to-end latency, one commercial approach is to build a proprietary on-machine computing system to map sensing and computing tasks to the best compute substrates to achieve optimal performance and energy consumption [3]. However, this design evolves through trial and error, and the whole design process takes much time with multiple iterations. This is the same approach that many autonomous machine companies take: they deploy *ad hoc* solutions to ensure on-time autonomous machine product release. These are product-specific and hard to generalize to other autonomous machine designs, hence leading to high re-engineering costs for each product.

**Challenge:** We conclude that the key technical challenge of designing autonomous machine compute system is to develop an appropriate computer architecture, along with a software stack that allows the flexibility of mapping various computation graphs from different types of autonomous machines to the same compute substrate, while meeting the real-time performance, cost, and energy constraints. Existing CPUs meet the flexibility requirement but fail to meet the performance and energy constraints, whereas other compute substrates, especially accelerators, typically target to meet the performance and energy constraints of one module, *e.g.,* Perception, without optimizing the end-to-end system.

## Area 2: The Sensing System

Autonomous machines gather information about their surroundings through their sensors; sensing is thus the very first stage of autonomous machines' deep processing pipeline. The ultimate objective of the sensing system is to provide accurate, reliable, and comprehensive information about the physical environment for later processing stages to precisely understand the physical world. These three goals (accuracy, reliability, and comprehensiveness) can be achieved through multi-modal sensor fusion [4]. However, multi-modal sensor fusion presents significant challenges because all sensor data have to be synchronized temporally and spatially all the while during the operation of the machine.

As an instance of temporal synchronization, take two sensors, a camera and a LiDAR; their measurements must be made simultaneously so that the machine can fuse the measurements and reconstruct an accurate and comprehensive view of the environment. Without proper time synchronization, the data from multiple sensors could yield an inaccurate, ambiguous view of the environment, leading to potentially catastrophic outcomes [5]. As for an example of spatial synchronization, assume multiple cameras with overlapping field of views are installed at different locations of the autonomous machine. In order to guarantee that the visual sensing results from these multiple cameras accurately match, results from each camera need to go through geometric transformations, or calibration, so that they can be projected to a commonly-agreed-upon reference point. Currently, on most autonomous machines that fuse multiple sensor modalities, the sensors must be manually calibrated [6], a slow, labor intensive, and often error prone process. Once this calibration has been completed it is often assumed that it remains unchanged while the autonomous machine is operating. In practice, the calibration is gradually degraded due to the autonomous machine's motion, leading to the need of periodic recalibration.

**Challenge:** In an ideal scenario, we should be able to simply plug a sensor into an autonomous machine, and it would just work. Unfortunately, we are still far from this ideal, and indeed any change to the sensor configuration of the autonomous machine must often lead to a redesign of the temporal and spatial synchronization subsystem (*e.g.,* a complete recalibration of the whole autonomous machine), or even a modification to the perception and localization systems. We conclude that the key technical challenge for the design of sensing systems is to provide a standardized framework for temporal and spatial synchronization for existing as well as new sensors. Only then will we be able to provide sensor plug-and-play capability for autonomous machines.

## Area 3: The Perception System

Perception is essential to any autonomous machine applications where sensory data and artificial intelligence techniques are involved. The final objective of perception is to extract spatial and semantic information from the raw sensing data so as to allow the machine to construct a comprehensive understanding of its operating environment. Such understanding includes the types, positions, headings, speeds, and dimensions of all objects in the environment.

There are two categories of perception for autonomous machines, the deep learning-based approach and the geometry-based approach. The *deep learning-based* approach is mainly used to extract semantic information and is heavily used in applications such as object detection, scene understanding, segmentation, tracking and prediction, *etc*. Within an autonomous machine's perception system, multiple deep learning models are running simultaneously, *e.g.,* networks for 2D perception, networks for 3D perception, and networks for tracking and prediction. Nonetheless, deep learning-based perception is often the performance bottleneck in autonomous machines. It thus becomes a painful tradeoff between perception quality and compute and energy resource utilization.

The *geometry-based* approach is mainly used to extract positional and dimensional information of the target objects. A most common instance of geometry-based perception application is the real-time stereo vision used for autonomous machine navigation, obstacle avoidance, and scene reconstruction. Stereo vision allows autonomous machines to obtain 3D structure information of the scene. A stereo vision system typically consists of two cameras to capture images from two points of view. Disparities between the corresponding pixels in two stereo images are detected using stereo matching algorithms. Depth information can then be calculated from the inverse of this disparity. Autonomous machines must fuse the positional and dimensional information and the semantic information of the target objects to form a comprehensive understanding of their environments.

**Challenge:** We conclude that the key technical challenge for perception is in the development of a general framework to generate reliable and precise understanding of the operating environment in real time. Reliable and precise perception can be achieved by fusing various perception results, such as 2D semantics, 3D semantics, and 3D geometry, an area still currently under active research. On the other hand, to achieve real-time performance, software approaches such as the compression-compiler co-design method that combines the compression of deep learning models and their compilation to optimize both the size and speed of deep learning models [7, 8]. Hardware approaches such as hardware accelerators for perception modules [9, 10] can be taken. Ultimately, more research is required to determine what combination of software and hardware approaches will be most effective in achieving real-time performance for the perception system.

## Area 4:  The Localization System

Fundamental to autonomous machines is localization, *i.e.,* ego-motion estimation, which calculates the position and orientation of an agent in each frame of reference. Formally, localization generates the six degrees of freedom (DoFs) pose, including the three DoFs for the translational pose, which specify the <x, y, z> position, and the three DoFs for the rotational pose, which specify the orientation about three perpendicular axes, *i.e.*, yaw, roll, and pitch. Knowing the translational pose fundamentally enables an autonomous machine to plan its path and to navigate, while the rotational pose further lets it stabilize itself.

Localization is highly sensitive to the operating environment, as different environments require different sensors and algorithms. For instance, in outdoor environments which usually provide stable GNSS signals, the compute-light visual inertial odometry (VIO) algorithm or LiDAR odometry coupled with GNSS signals achieves the best accuracy and performance. In contrast, in unknown, unmapped indoor environments, a LiDAR or visual Simultaneous Localization and Mapping (SLAM) algorithm delivers the best accuracy [11]. Nonetheless, different localization algorithms often incur different latencies, and even worse, latency variations.  For instance, for visual odometry or visual SLAM, the processing latency often depends on the number of feature points extracted from the current image, which means large latency variations that might impact predictability and safety.

**Challenge:**  We conclude that the key technical challenge in localization is to develop a standard framework capable of adapting to different operating scenarios by unifying core primitives in various localization algorithms, and to seamlessly switch between different algorithms as the autonomous machine navigates through different environments. In addition, this standard framework should provide a desirable software baseline for acceleration which will minimize processing latency as well as latency variations.

## Area 5:  The Planning and Control System

The planning and control system dictates how an autonomous machine should maneuver. Traditional planning and control systems must include behavioral decisions, motion planning and feedback control kernels [1].  Specifically, motion planning entails three steps, namely roadmap construction, collision detection, and graph search. As autonomous machines work with configurations with higher degrees of freedom (*e.g.,* robotic arms), motion planning will become increasingly complex since the search space will have exponentially increased.

While commonly deployed in commercial autonomous machines, traditional planning and control methods often utilize human-in-the-loop rule-based approaches, where engineers fine tune the planning and control kernels with available test data.  This approach is not only slow and costly, but also not robust as all "corner" cases need to be added manually.  As more rules must be included, the planning and control system becomes increasingly large and unwieldy and often fails to meet real-time requirements. In addition, rule-based methods suffer from a notorious difficulty to handle the multi-agent problem, or the fact that the actions taken by an agent can affect the behavior of other machines in the same environment, hence failing to handle complex traffic scenarios.

Recently, the Deep Reinforcement Learning (DRL) for planning and control is being actively researched in many places worldwide. Compared to traditional planning and control methods, inference with DRL incurs low computational requirements during operation, especially for high degree of freedom configurations [12]. In addition, DRL methods are capable of handling the multi-agent problem, allowing autonomous machines to handle complex traffic scenarios. However, model training is the bottleneck for DRL-based planning and control systems, as model training requires a vast number of trials to gain enough experience. This is especially the case for complex scenarios where model training can easily reach millions of steps, with each setup of hyper-parameters or reward hypothesis taking hours or even days.

**Challenge:** While DRL-based planning and control methods are highly promising, the key technical challenge for planning and control remains the development of a cloud infrastructure that provides sufficient compute power and generates high-quality data for DRL model training. First, to ensure we generate enough high-quality data to train the DRL networks, we need to develop simulation engines that are capable of closely simulating various physical scenarios. Second, to greatly improve algorithm development efficiency, especially for complex scenarios, we need to develop a model training infrastructure that can reduce the training time by orders of magnitudes.

## Area 6:  The Communication System

As most autonomous machines are also connected machines, the communication among autonomous machines, as well as the communication between autonomous machines and the infrastructure are critical to overall performance, safety, and reliability.  For instance, cellular Vehicle-to-Everything (C-V2X) is designed as a unified connectivity platform which provides low latency communications. It consists two modes of communications. The first mode uses direct communication links between vehicles, infrastructure, and pedestrians. The second mode leverages commercial LTE or 5G cellular networks to enable vehicles to receive information from the internet.  With the proliferation of 5G technologies, we expect that in the near future, the C-V2X technology will be extended to all autonomous machines [13].

Disappointingly, the field experimental results demonstrate that the real-world wireless networks are unstable and unreliable, which and can significantly affect the arrival timing of the infrastructure-side data [15]. Further, complex and ever-changing traffic conditions can exacerbate this situation. Specifically, while direct communication C-V2X networks, communication latency is less of a problem, the communication bandwidth remains insufficient to sustain the data needed for cooperative autonomous machines.  With commercial LTE or 5G networks, while bandwidth is less of a problem, latency variations often lead to missed deadlines where the miss ratio can be as high as 30%, thereby preventing reliable cooperative autonomous driving scenarios.

**Challenge:** The key technical challenge of designing autonomous machine communication system is to develop a highly reliable autonomous machine communication network with bandwidth guarantees and minimum latency variations. There are some initial directions of research that may achieve these goals, but more research is required: first, we can adjust the priority in the commercial communication networks to provide the highest priority to autonomous machines connections as they are safety-critical. Second, a virtual communication backbone that builds on top of both the LTE or 5G commercial networks as well as the direct communication C-V2X networks can improve the reliability of autonomous machine communication through intelligent scheduling. Third, we can develop intelligent fusion engines on autonomous machines to dynamically decide whether it is reliable and safe to fuse infrastructure-side data.

## Area 7: The Cloud System

Autonomous machines, especially the ones with mobility (*e.g.,* vehicles, delivery robots, service robots, drones), require cloud computing support. Such an autonomous machine cloud system must be capable of supporting offline computing tasks such as deep learning model training or high-definition (HD) map generation, or even online computing tasks such as aggregating all traffic information and broadcasting the information to all traffic participants, especially in the context of cooperative autonomous driving.

Autonomous machine clouds today are mostly application specific although they share many common requirements. To support different cloud applications, we need an infrastructure to provide basic services including distributed computing, distributed storage, as well as hardware acceleration through heterogeneous computing. If we were to tailor the infrastructure for each application, we would have to maintain multiple infrastructures, potentially leading to low resource utilization, low performance, and high management overhead. One effective way to solve this problem is to develop a unified autonomous machine cloud infrastructure with consolidated distributed computing and storage services [14].

Simulation is the cornerstone application in autonomous machine cloud systems, in which autonomous machine functions are exercised in a virtual environment. The architecture of the autonomous machine cloud simulation system is illustrated in Figure 3, the simulator models the sensing mechanism and generates sensory data by simulating interactions between sensors and environments. Exercised by the simulated sensory data, the autonomous machine software generates control commands to maneuver. We believe that autonomous machine clouds should be simulation-centric, as simulation consumes outputs from most other cloud-based applications, such as HD map generation and deep learning model training. Also, simulation consumes more than 80% of computing resources in our autonomous machine cloud, making it a low hanging fruit in terms of a target for optimization. In addition, by moving some of the physical testing to cloud-based simulation, we managed to reduce testing costs by 28x and to improve testing efficiency by 12x.

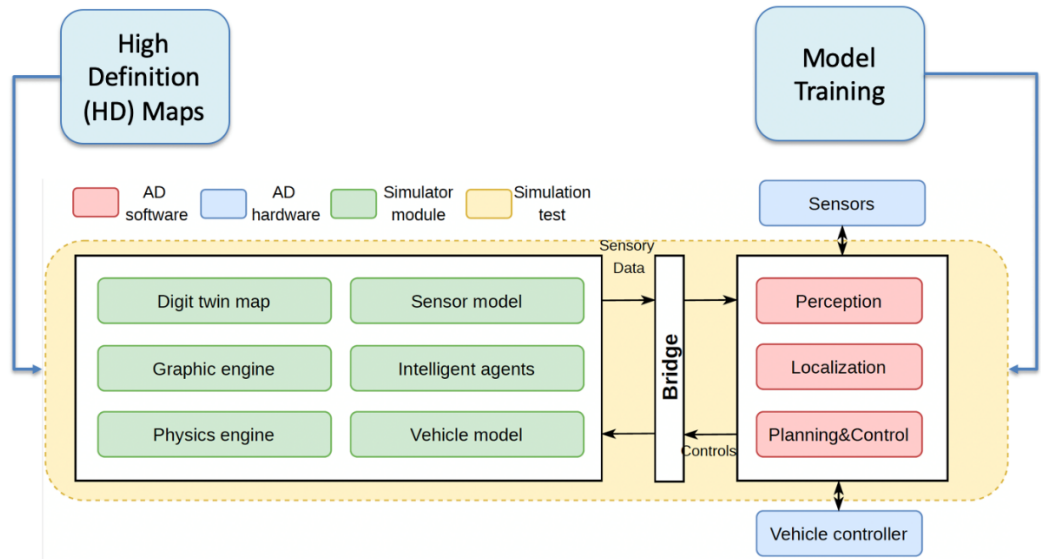

Figure 3: A cloud system for autonomous machines

**Challenge:** Moving forward, as autonomous machine clouds become more prevalent, two technical challenges have emerged: first, we need to continuously develop and refine distributed computing and storage services to adapt to the needs of various autonomous machine applications, especially to provide the infrastructure support to satisfy the ever-increasing compute and storage demands for autonomous machines. Second, we need to deploy and optimize basic simulation services on the cloud infrastructure to serve various autonomous machine simulation needs. For instance, autonomous drones and service robots demand different simulations services.

## Area 8: The Cooperation among Autonomous Machines

While traditional autonomous machines utilize only on-machine intelligence, cooperative autonomous machines depend on the cooperation between autonomous machines in addition to the infrastructure. Take autonomous driving for example, the infrastructure-vehicle cooperative autonomous driving approach relies on the cooperation between intelligent roads and intelligent vehicles. This approach is not only safer but also more economical compared to the traditional on-vehicle-only autonomous driving. Based on the progress towards commercial deployment, a three-stage development roadmap has been proposed as follows [15]:

- Stage 1: infrastructure-augmented autonomous driving (IAAD), in which autonomous vehicles fuse vehicle-side and infrastructure-side perception outputs to improve safety of autonomous driving.

- Stage 2: infrastructure-guided autonomous driving (IGAD), in which autonomous vehicles can offload all the proactive perception tasks to the infrastructure in order to reduce per-vehicle deployment costs.
- Stage 3: infrastructure-planned autonomous driving (IPAD), in which the infrastructure takes care of both perception and planning, thus achieving maximum traffic efficiency and cost efficiency.

To complete these tasks, Figure 4 presents an overview of the cooperative autonomous driving system architecture. It consists of the Systems on Vehicle (SoVs), the Systems on Road (SoRs), the intelligent transportation cloud system (ITCS), and the control center. The SoRs provide local perception results to the SoVs for blind spot elimination and extended perception to improve safety. Meanwhile, the SoRs process incoming sensor data and send the extracted semantic data to the ITCS for further processing. The ITCS fuses all incoming semantic data to generate global perception and planning information before the control center can dispatch real-time global traffic information, navigation plans, and even vehicle control commands to the SoVs to achieve optimal traffic efficiency. This design not only applies to autonomous driving but can be generalized to all autonomous machines.

**Challenge:** The key technical challenge for cooperative autonomous machines is integration, as its full realization relies on all the technical areas described in the previous sections. To guide the progress of cooperative autonomous machines, the community needs to clearly define the technical specifications and standards of each technical area for each stage of deployment so as to ensure effective technical integration.

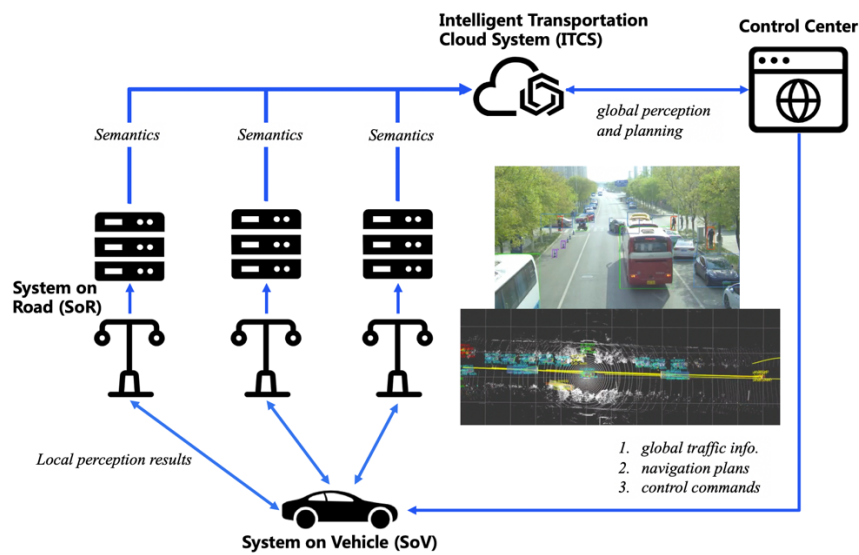

Figure 4: An infrastructure-vehicle cooperative autonomous driving system

## Area 9:  The Education of Engineers in the Age of Autonomous Machines

Talent supply is the single most important input for the ubiquitous deployment of autonomous machines.  Unfortunately, in the past few years, industry has observed a huge supply-demand gap for autonomous machine engineers.

On the supply side, many students actually have a high degree of interest in autonomous machine technologies but find that their home universities do not provide an adequate education to prepare them to enter this field. On the demand side, it is extremely difficult for autonomous machine companies to hire qualified engineers to fill the open positions, thus forcing many companies to develop internal training programs to prepare incoming engineers. However, since many companies in this field are resource-constrained startups, the need to provide training for incoming engineers imposes additional burden to their financial situation, leading to inefficient allocation of precious funding resources.

To bridge this supply-demand gap, we as a society, should urgently reshape our engineering education programs and create a cross-disciplinary program to impart students with a technical background in computer science, computer engineering, electrical engineering, as well as mechanical engineering [16]. On top of this cross-disciplinary technical foundation, a capstone project that will provide students with hands-on experience of working with a real autonomous machine would consolidate the technical foundation [17].  An effective way to perform the capstone project is through cooperative education (co-op), a structured method combining classroom-based education with practical work experience which would require close collaboration between universities and industry.

**Challenge:**  This proposed engineering education reform roadmap is ambitious and we have a long way to go to achieve it, as it is extremely challenging to establish long-term and stable collaboration among multiple parties. To promote the collaboration between industry and academia in emerging fields like autonomous machines, we advocate for governments to provide incentives through grants programs.  It is notable that the U.S. government has been playing an active in driving the U.S. innovation rate [18].  These education and training grants should advance the domestic workforce skills in autonomous machine technologies, and consequently improve the domestic economy.

## Area 10:  The Broader Societal Impact of Autonomous Machines

Finally, autonomous machines are expected to completely revolutionize our economy by greatly improving the efficiency in delivery, transportation, manufacturing, and many other sectors. Take last-mile autonomous delivery for example, the growing labor cost may be prohibitive for service providers: in China, a contracted delivery clerk with an annual salary of $20,000 can deliver 110 parcels per day, which means that each delivery order costs nearly $0.5. This cost is expected to continue increasing as the demographic dividend has reached its end. Deploying autonomous machines for last-mile delivery has been proven to be a promising approach to cut these costs by more than half [19].

Take autonomous vehicles for another example, traffic inefficiency such as congestion imposes high costs on our society. For the trucking industry alone, the American Transportation Research Institute estimates that congestion costs the U.S. $74.1 billion annually, of which $66.1 billion occurs in urban areas. As one of the ultimate goals of autonomous driving is to completely eliminate traffic inefficiency, by utilizing autonomous driving technologies on trucking alone can lead to a significantly more efficient national economy.

However, the benefits of improved efficiency do come at a cost. Converting all vehicles in the world into autonomous vehicles would result in an enormous social cost, as autonomous vehicles are still very expensive to build [20]. An alternative but cost-efficient solution is cooperative autonomous machines. As the infrastructures become more intelligent, more workloads can be offloaded to the infrastructure side, greatly reducing the hardware and energy costs of autonomous machine deployments. It is estimated that this approach would drop the cost of autonomous machine deployment by more than 50% [15]. Besides the impact on the economy, there will also be increasingly more ethical, privacy, and security challenges as autonomous machines proliferate.

**Challenge:** One non-technical but critical challenge is for the whole industry to thoroughly understand the potential impact of various autonomous machines on our economy, as well as the costs of large-scale deployments of various autonomous machines. Only with accurate projections and estimations will we be able to provide comprehensive information to policy makers and technology firms. Consequently, policy makers and technology firms are capable of making their policy and technology investment decisions with the ultimate goal of improving human society with autonomous machine technologies. Once we have a clear understanding of the societal benefits of autonomous machines, our society shall develop ethical guidelines, standards, and legislations to guide the healthy development of autonomous machines. Hence, an immediate next step is for economists, policy makers, and autonomous machine engineers to work closely together to define economic impact and policy roadmaps to integrate autonomous machines into our daily life.

## Summary

After more than six decades of information technology development, we believe that autonomous machines will completely revolutionize our daily life and our economy in the coming decade. The impact of autonomous machines on our society is likely to be much deeper and broader than any other information technology revolution that we have experienced in the past decades. To facilitate the rise of autonomous machines, in this article we have categorized the technical and non-technical challenges in ten areas; and for each area, we have reviewed the current status, roadblocks, and potential research directions. Note that the current categorization is by no means comprehensive, as we expect new research areas will emerge as the field evolves. Rather, this article is meant to serve as an initial step for our community to define clear and effective roadmaps to make the age of autonomous machines a reality in the near future. As a next step, we aim to work with technical, educational, economic, and policy experts, from academia, industry, and government, within each area to form cross-disciplinary teams to address the challenges laid out in this paper.